\documentclass{article} 
\usepackage{iclr2017_conference,times}
\usepackage{hyperref}
\hypersetup{
    colorlinks = true,
    linkcolor = {blue},
    citecolor = {magenta},
    urlcolor = {green}
}
\usepackage{url}
\usepackage[pdftex]{graphicx}
\usepackage{wrapfig}
\usepackage{booktabs}       
\usepackage{subfig}
\usepackage{amsmath}
\usepackage{float}
\usepackage{amssymb}
\usepackage[labelfont=bf,textfont={},font=small,justification=justified,indention=0cm]{caption}

\usepackage{tikz}
\usetikzlibrary{positioning,shapes,arrows,fit}

\title{Changing Model Behavior at Test-Time Using Reinforcement Learning}

\author{Augustus Odena\thanks{Equal Contribution} \hspace{0.01cm} \thanks{Work completed
    as a participant in the 2016-2017 Google Brain Residency program.}
, Dieterich Lawson\footnotemark[1]  \hspace{0.01cm} \footnotemark[2]
, Christopher Olah
\\
Google Brain\\
\texttt{\{augustusodena,dieterichl,colah\}@google.com} \\
}

\begin{document}

\tikzset{%
  block/.style    = {draw, thick, rectangle, minimum height = 3em,
    minimum width = 3em},
  sum/.style      = {draw, circle, node distance = 2cm}, 
  input/.style    = {coordinate}, 
  output/.style   = {coordinate} 
}

\def\reals{\mathbb{R}}

\maketitle

\section{Introduction}

Machine learning models are often used at test-time
subject to constraints and trade-offs not present at training-time.
A computer vision model operating on an embedded device may
need to perform real-time inference; a
translation model operating on a cell phone may wish to bound
its average compute time in order to be power-efficient.
In these cases, there is often a tension between
satisfying the constraint and achieving acceptable model performance.
These constraints need not be restricted to speed and accuracy,
but can reflect preferences for model simplicity or other
desiderata.

One way to deal with constraints is to build them into models
explicitly at training time.
This has two major disadvantages:
First, it requires manually designing and retraining a new model for each use case.
Second, it doesn't permit adjusting constraints at test-time in an input-dependent way.

In this work, we describe a method to change model behavior at test-time on a per-input basis.
This method involves two components:
The first is a model that we call a Composer.
A Composer adaptively constructs computation graphs from sub-modules on a per-input basis
using a controller trained with reinforcement learning to examine intermediate activations.
The second is the notion of Policy Preferences, which allow test-time modifications
of the controller policy.

This technique has several benefits:
First, it allows for dynamically adjusting for constraints at inference 
time with a single trained model.
Second, the Composer model can `smartly' adjust for constraints
in the way that is best for model performance (e.g. it can
decide to use less computation on simpler inputs).
Finally, it provides some interpretability, since we can see which examples
used which resources at test-time.

The rest of this paper is structured as follows:
In Section \ref{section:composer} we describe the Composer model,
in Section \ref{section:preferences} we describe Policy Preferences, and
in Section \ref{section:experiments} we present experimental results.

\section{The Composer Model} \label{section:composer}

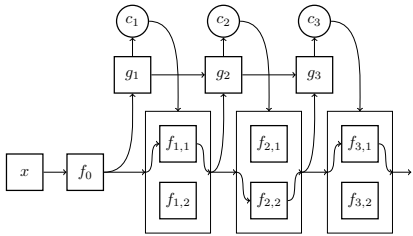
\begin{figure}[!h]
  \vspace{-0.2cm}
\centering
\resizebox{0.4\textwidth}{!}{%
\begin{tikzpicture}[
square/.style={rectangle, draw=black,  thick, minimum height=8mm, minimum width=8mm},
round/.style={circle, draw=black, thick, minimum size=7mm},
]
\node [square] (image) {$x$};
\node [square] (stem) [right=0.5cm of image] {$f_0$};
\node (middle_1) [right=0.5cm of stem] {};
\node  (network1_2) [right=0.75cm of middle_1] {};
\node [square] (network1_1) [above=0.085cm of network1_2] {$f_{1,1}$};
\node [square] (network1_3) [below=0.085cm of network1_2] {$f_{1,2}$};

\node[draw=black, fit=(network1_1) (network1_2) (network1_3), inner sep=3mm] (network1) {};

\node (middle_2) [right=0.75cm of network1_2] {};

\node [] (network2_2) [right=0.75cm of middle_2] {};
\node [square] (network2_1) [above=0.085cm of network2_2] {$f_{2,1}$};
\node [square] (network2_3) [below=0.085cm of network2_2] {$f_{2,2}$};

\node[draw=black, fit=(network2_1) (network2_2) (network2_3), inner sep=3mm] (network2) {};

\node (middle_3) [right=0.75cm of network2_2] {};

\node [] (network3_2) [right=0.75cm of middle_3] {};
\node [square] (network3_1) [above=0.085cm of network3_2] {$f_{3,1}$};
\node [square] (network3_3) [below=0.085cm of network3_2] {$f_{3,2}$};
 
 \node[draw=black, fit=(network3_1) (network3_2) (network3_3), inner sep=3mm] (network3) {};
 
\node [square] (controller_1) [above=1.6cm of middle_1] {$g_1$};
\node [square] (controller_2) [above=1.6cm of middle_2] {$g_2$};
\node [square] (controller_3) [above=1.6cm of middle_3] {$g_3$};

\node [round] (c_1) [above=0.4cm of controller_1] {$c_1$};
\node [round] (c_2) [above=0.4cm of controller_2] {$c_2$};
\node [round] (c_3) [above=0.4cm of controller_3] {$c_3$};

\node (output) [right=1cm of network3_2] {};
\draw[->] (image.east) -- (stem.west);
\draw[->] (stem.east) -- (network1.west);
\draw[->] (stem.east) to[out=0, in=-90, looseness=1] (controller_1.south);
\draw[->] (controller_1.north) -- (c_1.south);
\draw[->] (c_1.east) to[out=0, in=90, looseness=0.9] (network1.north);
\draw[->] (network1.west) to[out=0, in=-180, looseness=1] (network1_1.west);
\draw[->] (network1_1.east) to[out=0, in=-180, looseness=1] (network1.east);

\draw[->] (network1.east) -- (network2.west);
\draw[->] (network1.east) to[out=0, in=-90, looseness=0.5] (controller_2.south);
\draw[->] (controller_2.north) -- (c_2.south);
\draw[->] (c_2.east) to[out=0, in=90, looseness=0.9] (network2.north);

\draw[->] (network2.west) to[out=0, in=-180, looseness=1] (network2_3.west);
\draw[->] (network2_3.east) to[out=0, in=-180, looseness=1] (network2.east);

\draw[->] (network2.east) -- (network3.west);
\draw[->] (network2.east) to[out=0, in=-90, looseness=0.5]  (controller_3.south);
\draw[->] (controller_3.north) -- (c_3.south);
\draw[->] (c_3.east) to[out=0, in=90, looseness=0.9] (network3.north);

\draw[->] (network3.west) to[out=0, in=-180, looseness=1] (network3_1.west);
\draw[->] (network3_1.east) to[out=0, in=-180, looseness=1] (network3.east);

\draw[->] (controller_1.east) -- (controller_2.west);
\draw[->] (controller_2.east) -- (controller_3.west);

\draw[->] (network3.east) -- (output);
\end{tikzpicture}
}
\caption{A diagram of the Composer model. This Composer 
has 3 metalayers each consisting of 2 modules. Circles denote stochastic nodes;
squares denote deterministic nodes.}
\label{fig:composer}
\vspace{-.4cm}
\end{figure}

The Composer Model (depicted in Figure \ref{fig:composer}) consists of a
set of modules and a controller
network. The modules are neural networks and are organized 
into `metalayers'. At each metalayer in the network, the 
controller selects which module will be applied to the activations from the 
previous metalayer (see also
\cite{FACTOREDREPRESENTATIONS,bengio2015conditional,QUESTIONANSWERING,OUTRAGEOUSLYLARGE,PATHNET,
DSNN}).

Specifically, let the $i$th metalayer of the network be denoted $f_i$, with 
$f_0$ a special book-keeping layer, called the `stem'. For $i>0$, the $i$th 
metalayer is composed of $m_i$ functions, $f_{i,1},\ldots, f_{i,m_i}$ that 
represent the individual modules in the metalayer. Note that these 
modules can differ in terms of 
architecture, number of parameters, or other characteristics. 
There are $n$ metalayers, so the depth of the network is $n+1$ (including the stem). 
Once a selection of modules is made, it defines a neural network,
which is trained with SGD.

The controller is composed of $n$ functions, $g_1,\ldots g_n$ each of which 
output a policy distribution over the $m$ modules in the 
corresponding metalayer. In equations:

\begin{align}
\vspace{-1cm}
    a_0 &= f_0(x)\\
    p_i &= g_i(a_{i-1}, c_{i-1})\\
    c_i &\sim \text{Categorical}(p_i)\\
    a_i &= f_{i,c_i}(a_{i-1})\\
    p(y|x,c_{1:n}) &= \text{softmax}(a_n)
\vspace{-1cm}
\end{align}

where $x$ is the input, $y$ is the ground truth labeling associated with $x$,
$a_i$ are the network activations after the $i$th metalayer, 
$p_i$ are the parameters of the probability distribution output by the 
controller at step $i$, $c_i$ is the module choice made by sampling from 
the controller's distribution at step $i$, $c_{1:n}$ denotes $c_1, \ldots, c_n$,
and $p_{1:n}$ similarly denotes $p_1, \ldots, p_n$.

The controller can be implemented as an RNN by making
the controller functions share weights and a hidden state.
We train the controller to optimize the reward $\log p(y|x,c_{1:n})$ using REINFORCE \citep{williams1992simple}.
For details, see Appendix \ref{appendix:trainingsetup}.

\section{Policy Preferences} \label{section:preferences}

We can augment the reward function of the controller to express preferences
we have about its policy over computation graphs.
Implemented naively, this does not allow for test-time modification of the preferences.
Instead, we add a cost $C(p_{1:n}, c_{1:n}, \gamma)$ to the reward for the controller,
where $\gamma$ is a preference value that can be changed on either a per-input 
or per-mini-batch basis.

Crucially, $\gamma$ is given as an input to the controller.
At train-time, we sample $\gamma \sim \Gamma$,
where $\Gamma$ is a distribution over preferences.
Because of this, the controller learns at training
time to modify its policy for different settings of $\gamma$.
At test time, $\gamma$ can be changed to correspond to changing preferences.
The specific dependence of $C$ on the action and the preference can vary:
$\gamma$ need not be a single scalar value.

Below we describe two instances of Policy Preferences applied to the Composer model.
Policy Preferences could also be applied in a more general reinforcement learning setting.

\subsection{Glimpse Preferences}

For a variety of reasons we might want
to express test-time preferences about model resource consumption 
(see also \cite{ACT, SACT}).
One environment where this might be relevant is when
the model is allowed to take glimpses of the input
that vary both in costliness and usefulness.
If taking larger glimpses requires using more parameters,
we can achieve this by setting

\begin{equation}
  C_1(p_{1:n}, c_{1:n}, \gamma) = -\gamma \sum_{i=1}^n \beta_{i,c_i}.
\end{equation}

where $\beta_{i} \in \reals^m$ is the vector of parameter counts for each module
in the $i$th metalayer and $\beta_{i,c_i}$ is the parameter count for
the module chosen at the $i$th metalayer. 
$C_1$ is applied per-element here.

\subsection{Entropy Preferences}

Because the modules and the controller are trained jointly,
there is some risk that the controller will give all of its
probability mass to the module that happens to have been trained the most so far.
To counteract this tendency we can use another cost

\begin{equation}
  C_2(p_{1:n}, c_{1:n}, \gamma) = -\gamma \sum_{i=1}^n \left\vert\left\vert\frac{1}{N}\sum_{j=1}^N p_{i}^{(j)}\right\vert\right\vert_2^2.
\end{equation}

where $N$ is the number of examples in a batch
and $p_i^{(j)} \in \reals^m$
is the vector of module probabilities 
produced by the controller at metalayer $i$ for batch element $j$.
Note that this reward is maximized when the controller
utilizes all modules equally within a batch.
$C_2$ is applied per-batch.

One could simply augment the controller reward with a similar bonus
and anneal the bonus to zero during training, but our method has
at least two advantages: First, if load balancing between specialized modules
must be done at test time, using Policy Preferences allows it to be done in a
`smart' way. Second, this method removes the need to search for and follow
an annealing schedule - one can stop training at any time and
set the test-time batch-entropy-preference to 0.

\section{Experimental Results} \label{section:experiments}

We test the claim that one can dynamically adjust the amount of computation at inference time
with a single trained model using a Composer and Policy Preferences.
We introduce a modified version of MNIST called Wide-MNIST to accomplish this:
Images in Wide-MNIST have shape $28 \times 56$, with a digit appearing in either
the left half or the right half of the image.
Each image is labeled with one of 20 classes (10 for `left' digits and 10 for `right' digits).
We train a Composer model with two modules - a large module that 
glimpses the whole input and a small module that glimpses only the left side.
The Composer is trained using both entropy preferences and glimpse preferences.
See Figure \ref{fig:arrow} for results, which support the above claim.
See Appendix \ref{appendix:qualitative} for qualitative experiments.

\begin{figure}[!t]
\hspace{-0.0\textwidth} 
\begin{center}
\includegraphics[width=0.8\textwidth]{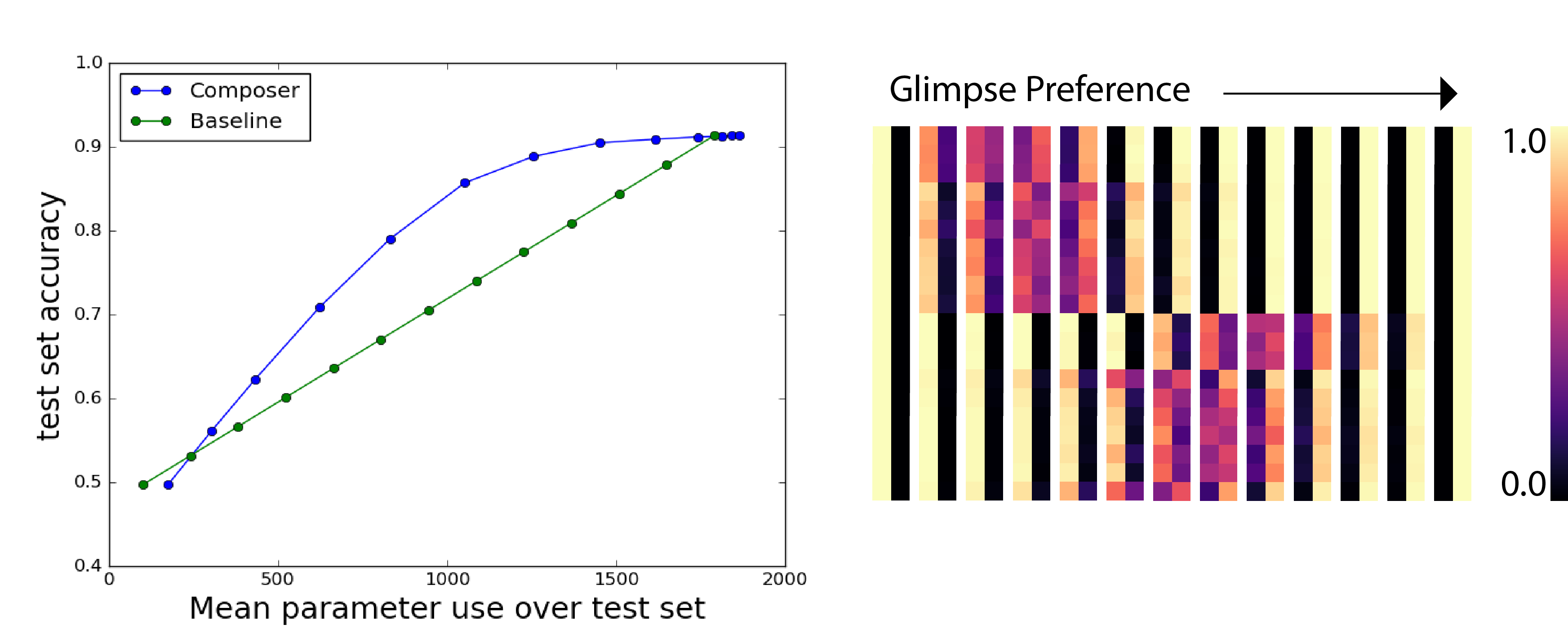}
\end{center}
\vspace{-5pt}
\caption{
 Changing the glimpse preference at test time for a model trained on the Wide-MNIST data-set.
 Note that all data points on the curve labeled `Composer' come from a \textbf{\textit{single}}
 trained model.
 (Left): The Composer model consists of a small module that only glimpses the left side of
 the input and a large model that glimpses the whole input, but uses more parameters to do so.
 It is evaluated at a variety of glimpse preference values,
 resulting in a variety of average-parameter-use values.
 A baseline model is created by randomly choosing between the large module
 and the small module some varying fraction of the time,
 in order to induce different average-parameter-use values.
 The gap between the Composer curve and the baseline curve represents the extent
 to which the controller has `smartly' adapted its module use
 based on the input example.
 (Right): Heat-maps denote probabilities assigned by the controller to each module,
 for each of the 20 labels, averaged over the whole test set.
 The left column corresponds to the small module and the right
 column to the large module.
 The top 10 rows correspond to `right' digits and the bottom 10
 rows correspond to `left' digits.
 As the glimpse preference is decreased, the controller first
 chooses to use the small module for `left' digits, since this
 doesn't cause much loss in classification accuracy.
 Only when the glimpse preference gets very low does the controller
 assign `right' digits to the small module.
}
\label{fig:arrow}
\vspace{-.4cm}
\end{figure}

\newpage

\section{Acknowledgements}
We thank Kevin Swersky, Hugo Larochelle, and Sam Schoenholz for helpful discussions.
\bibliography{paper}
\bibliographystyle{iclr2017_conference}

\newpage

\appendix

\section{Policy Gradient Details} 
\label{appendix:trainingsetup}
For simplicity let $c = c_{1:n}$ and $p = p_{1:n}$. Then, we use REINFORCE/policy gradients 
to optimize the objective
 \begin{equation}
   \textbf{E}_{c}\left[\log p(y|x,c) \right]
 \end{equation}
Jensen's inequality tells us that this is a lower bound on what we truly seek to optimize, $\log p(y|x)$
\begin{equation*}
  \log p(y|x) = \log \textbf{E}_{c}\left[p(y|x,c)\right] \geq \textbf{E}_{c}\left[\log p(y|x,c) \right].
 \end{equation*}
To maximize this expectation we perform gradient ascent. The gradient of this 
quantity with respect to the model parameters $\theta$ is
\begin{equation}
  \textbf{E}_{c}\left[ \nabla_{\theta} \log p(y|x,c) + \log p(y|x,c) \nabla_{\theta} \log p(c|x) \right].
\end{equation}
To implement the policy preference cost functions we can modify our reward to 
include $C(p, c, \gamma)$. In this case our objective becomes
 \begin{equation}
   \textbf{E}_{c}\left[\log p(y|x,c) + C(p,c,\gamma)\right]
 \end{equation}
 and the gradient becomes
\begin{equation}
  \textbf{E}_{c}\left[ \nabla_{\theta}(\log p(y|x,c) + C(p,c,\gamma)) + (\log p(y|x,c) + C(p,c,\gamma)) \nabla_{\theta} \log p(c|x) \right].
\end{equation}

\section{Qualitative Experiments}
\label{appendix:qualitative}

This section contains various qualitative explorations of Composer models
trained with and without Policy Preferences.

\subsection{Module Specialization}

\begin{figure}[H]
\hspace{-0.0\textwidth} 
\begin{center}
\includegraphics[width=1.0\textwidth]{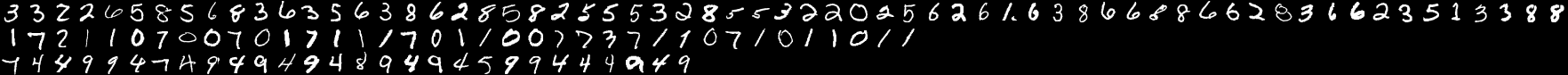}%
\end{center}
\vspace{-5pt}
\caption{Modules choices in the first metalayer of a random minibatch
  for a Composer with 3 modules at each metalayer.
  The number of parameters in each module increases from top to bottom,
  with the second module having twice as many parameters as the first,
  and the third module having three times as many.
  This Composer was trained using a Policy Preference penalizing the
  parameter count of the chosen module.
  Each row corresponds to a module,
  and the module size increases from top to bottom.
  The bottom module seems to act as a 4,7,9 disambiguator.
}
\label{fig:mnist_heatmap}
\end{figure}

On MNIST, modules often specialize to digit values of of 8, 5, and 3 or 4, 7, and 9,
which are the most frequently confused groups of digits.
See Figure \ref{fig:mnist_heatmap} for an example of this behavior.

\begin{figure}[H]
\hspace{-0.0\textwidth} 
\begin{center}
\includegraphics[width=0.9\textwidth]{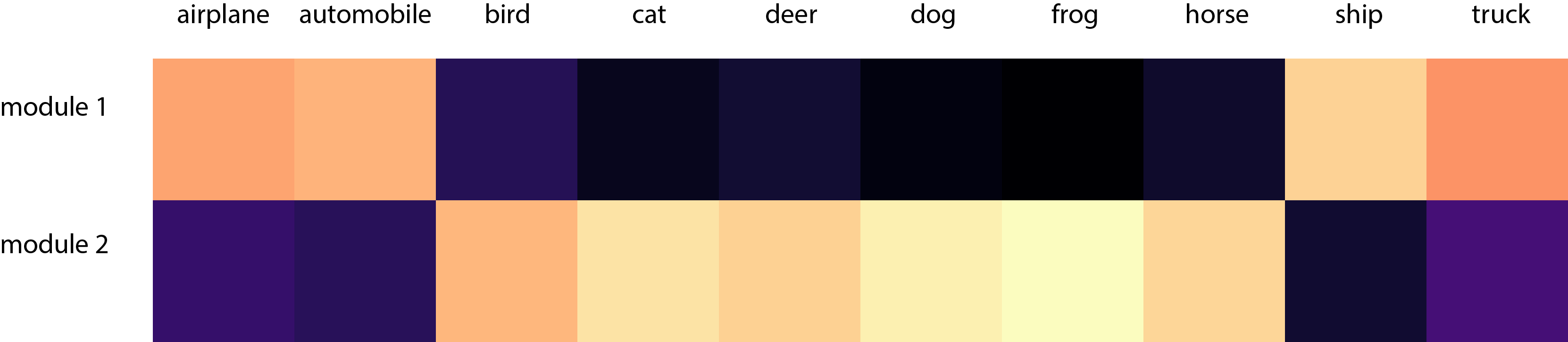}%
\end{center}
\vspace{-5pt}
\caption{A heatmap demonstrating the probability the controller assigned
  to using each module in the first metalayer, for each class in CIFAR-10.
  Note that the controller has divided up the data into essentially
  two groups - man-made things and natural things.
}
\label{fig:cifar_heatmap}
\end{figure}

Tests on CIFAR-10 yielded similar results:
See Figure \ref{fig:cifar_heatmap}.

\subsection{Entropy Preferences}

\begin{figure}[H]
\hspace{-0.0\textwidth} 
\begin{center}
\includegraphics[width=0.25\textwidth]{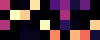}%
\hspace{10pt}
\includegraphics[width=0.25\textwidth]{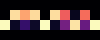}%
\end{center}
\vspace{-5pt}
\caption{The effect of the batch entropy preference on module selection frequency for
  a Composer with 1 metalayer and 4 modules trained on MNIST.
  These are module selection heatmaps similar to those in the above figures.
  (Left) With the entropy preference, modules are utilized more equally.
  After 100k steps, the mutual information between module choices and class labels
  for this run is 0.9 nats.
  (Right) With a normal entropy penalty, some modules `get ahead' and others
  never catch up.
  After 100k steps, the mutual information for this run is only 0.43 nats.
  This test was conducted with a constant, zero-variance value for the
  preference, so in this case the non-batch entropy preference is
  equivalent to the standard entropy penalty (which involves a
  sum of separate entropy penalties - one per example).
}
\label{fig:batch_bonus}
\end{figure}

\begin{figure}[H]
\hspace{-0.0\textwidth} 
\begin{center}
\includegraphics[width=0.25\textwidth]{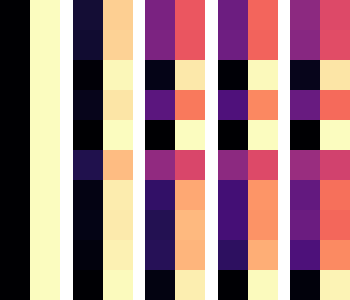}%
\end{center}
\vspace{-5pt}
\caption{
  Modifying the test-time entropy preference of a model trained
  using an `unbatched' entropy penalty.
}
\label{fig:entropy}
\end{figure}

We also conducted an ablation experiment demonstrating that
the batch entropy preference results in better module utilization.
See Figure \ref{fig:batch_bonus} for results.
In addition, we show the result of modifying the test-time
entropy preference of a model that has been trained with a standard
entropy penalty (Figure \ref{fig:entropy}). This could be useful in a variety of contexts
(e.g. language modeling).

\end{document}